\documentclass[10pt,twocolumn,letterpaper]{article}

\usepackage{wacv}      % To produce the REVIEW version for the algorithms track
\usepackage{graphicx}
\usepackage{amsmath}
\usepackage{amssymb}
\usepackage{booktabs}

\usepackage{subcaption}
\usepackage{xcolor}

\usepackage{bbm}
\usepackage{mathrsfs}
\usepackage{color, colortbl}
\definecolor{Gray}{gray}{0.9}
\definecolor{LightCyan}{rgb}{0.88,1,1}
\newcolumntype{a}{>{\columncolor{LightCyan}}c}

\usepackage[pagebackref,breaklinks,colorlinks]{hyperref}

\usepackage[capitalize]{cleveref}
\crefname{section}{Sec.}{Secs.}
\Crefname{section}{Section}{Sections}
\Crefname{table}{Table}{Tables}
\crefname{table}{Tab.}{Tabs.}

\def\wacvPaperID{130} % *** Enter the WACV Paper ID here
\def\confName{WACV}
\def\confYear{2024}

\begin{document}

\title{FG-Net: Facial Action Unit Detection with Generalizable Pyramidal Features}

\author{Yufeng Yin$^{1}$\thanks{Work done during internship at ByteDance.}, Di Chang$^{1}$, Guoxian Song$^{2}$, Shen Sang$^{2}$, Tiancheng Zhi$^{2}$, \\
Jing Liu$^{2}$, Linjie Luo$^{2}$, Mohammad Soleymani$^{1}$ \\
$^1$ University of Southern California, $^2$ ByteDance \\
\{yufengy, dichang, msoleyma\}@usc.edu, \\
\{guoxiansong, shen.sang, tiancheng.zhi, jing.liu, linjie.luo\}@bytedance.com}
\maketitle

%%%%%%%%% ABSTRACT
\begin{abstract}
Automatic detection of facial Action Units (AUs) allows for objective facial expression analysis. Due to the high cost of AU labeling and the limited size of existing benchmarks, previous AU detection methods tend to overfit the dataset, resulting in a significant performance loss when evaluated across corpora. To address this problem, we propose \textbf{FG-Net} for generalizable facial action unit detection.
Specifically, FG-Net extracts feature maps from a StyleGAN2 model pre-trained on a large and diverse face image dataset. Then, these features are used to detect AUs with a Pyramid CNN Interpreter, making the training efficient and capturing essential local features. The proposed FG-Net achieves a strong generalization ability for heatmap-based AU detection thanks to the generalizable and semantic-rich features extracted from the pre-trained generative model. Extensive experiments are conducted to evaluate within- and cross-corpus AU detection with the widely-used DISFA and BP4D datasets. Compared with the state-of-the-art, the proposed method achieves superior cross-domain performance while maintaining competitive within-domain performance. In addition, FG-Net is data-efficient and achieves competitive performance even when trained on 1000 samples. Our code will be released at \url{https://github.com/ihp-lab/FG-Net}
\end{abstract}

%%%%%%%%% BODY TEXT
\section{Introduction}
\begin{figure}[t]
    \centering
    \includegraphics[trim=0 0 0 0, clip, width=0.4\textwidth]{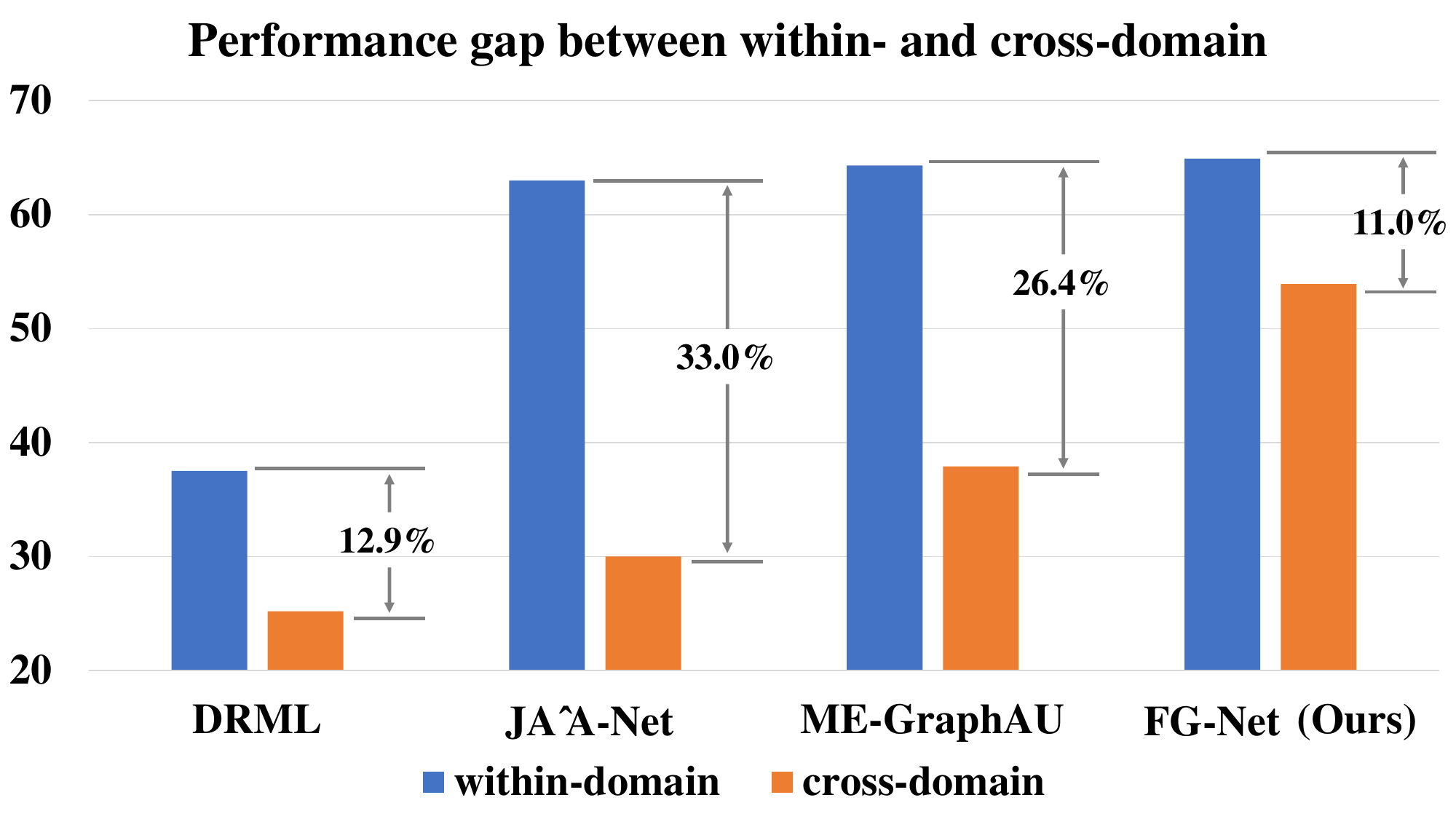}
    \caption{Performance (F1 score $\uparrow$) gap between the within- and cross-domain AU detection for DRML \cite{zhao2016deep}, J$\hat{\text{A}}$A-Net \cite{shao2021jaa}, ME-GraphAU \cite{luo2022learning}, and the proposed FG-Net. The within-domain performance is averaged between DISFA and BP4D, while the cross-domain performance is averaged between BP4D to DISFA and DISFA to BP4D. The proposed FG-Net has the highest cross-domain performance, thus, superior generalization ability.}
    \label{fig:front}
\end{figure}

Automatic detection of facial action units is a fundamental block for objective facial expression analysis \cite{ekman1977facial}. 
Manual annotations for facial action units are cumbersome and costly, as they require trained coders to label each frame individually. Common AU datasets, \ie, DISFA \cite{mavadati2013disfa} and BP4D \cite{zhang2014bp4d}, only contain a limited number of subjects (27 and 41 subjects respectively). Recent methods for AU detection \cite{zhao2016deep, shao2021jaa, luo2022learning} focus on deep representation learning, requiring a large number of samples.  
Existing AU detection methods are often evaluated with within-domain cross-validation, with training and testing data from the same dataset, and the generalization to other datasets (model trained and tested on different datasets) has not been widely investigated.
As within-domain performance can be due to \textbf{overfitting}, cross-corpus performance can suffer \textbf{a considerable loss} (see comparisons in Figure \ref{fig:front}).

In the field of semantic segmentation, recent studies \cite{zhang2021datasetgan, baranchuk2021label} leverage a well-trained generative model to synthesize image-annotation pairs from only a few labeled examples (around 30 training samples). They show that the intermediate features of generative models exhibit semantic-rich representations that are well-suited for pixel-wise segmentation tasks in a few-shot manner. Li \etal \cite{li2021semantic} showcase extreme out-of-domain generalization ability from such approaches. However, to the best of our knowledge, no existing work adapts such architectures to AU detection, potentially due to the following limitations: (i) the high dimensionality of the pixel-wise features results in inefficient training, and (ii) inference with pixel-level features lacks the information from the nearby regions, which is crucial to AU detection \cite{zhao2016deep, shao2021jaa}.

\begin{figure*}[t]
    \centering
    \includegraphics[width=0.8\textwidth]{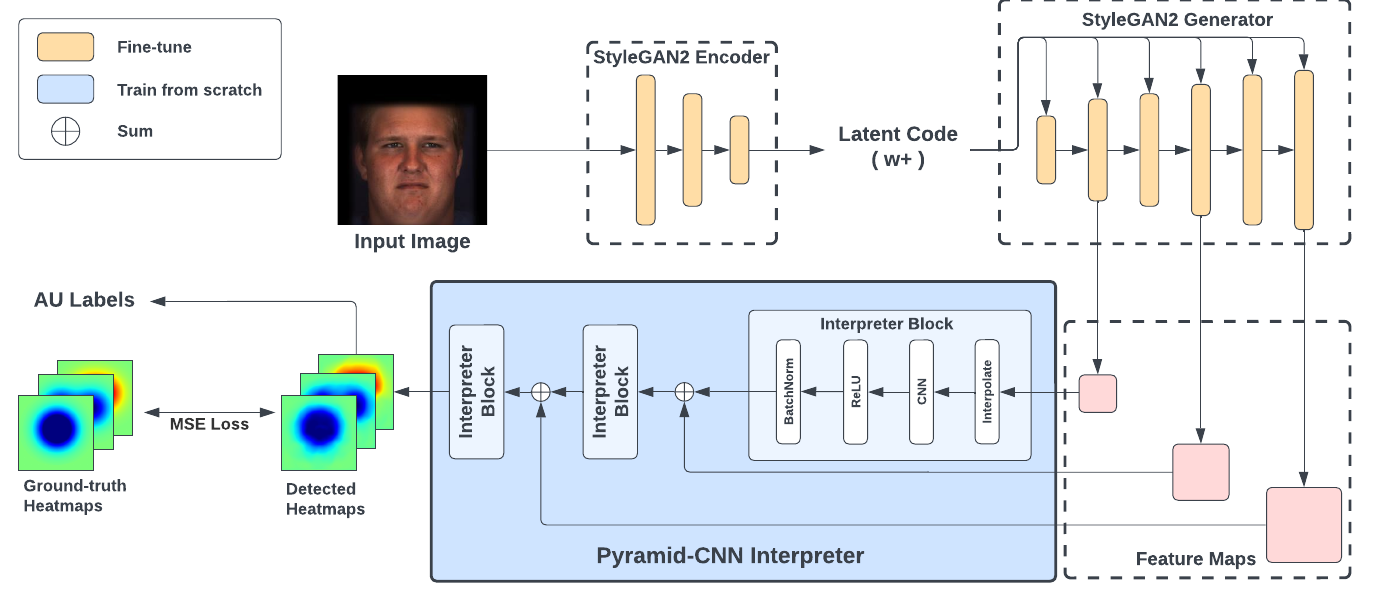}
    \caption{Overview of our proposed pipeline. FG-Net first encodes the input image into a latent code using a StyleGAN2 encoder (e.g. pSp \cite{richardson2021encoding} here). In the decoding stage \cite{karras2020analyzing}, we extract the intermediate multi-resolution feature maps and pass them through our Pyramid CNN Interpreter to detect AUs coded in the form of heatmaps. Mean Squared Error (MSE) loss is used for optimization between the ground truth and predicted heatmaps.}
    \label{fig:model}
\end{figure*}

In this paper, inspired by the success of GAN features in semantic segmentation, we propose \textbf{FG-Net}, a facial action unit detection method that can better generalize across domains. The general idea of FG-Net is to extract the generalizable and semantic-rich deep representations from a well-trained generative model (see Figure \ref{fig:model}). Specifically, FG-Net first encodes and decodes the input image with a StyleGAN2 encoder (pSp) \cite{richardson2021encoding}, and a StyleGAN2 generator \cite{karras2020analyzing}, trained on the FFHQ dataset \cite{karras2019style}. Then, FG-Net extracts feature maps from the generator during decoding. To take advantage of the informative pixel-wise representations from the generator, FG-Net detects the AUs through heatmap regression. We propose a Pyramid CNN Interpreter which incorporates the multi-resolution feature maps in a hierarchical manner. The proposed module makes the training efficient and captures essential information from nearby regions. Thanks to the powerful features from the generative model pre-trained on a large and diverse facial image dataset, the proposed FG-Net obtains a strong generalization ability and data efficiency for AU detection.

To demonstrate the effectiveness of our proposed method, we conduct extensive experiments with the widely-used DISFA \cite{mavadati2013disfa} and BP4D \cite{zhang2014bp4d} for AU detection. The results show that the proposed FG-Net method has a strong generalization ability and achieves state-of-the-art cross-domain performance (see Figure \ref{fig:front}). In addition, FG-Net achieves comparable or superior within-domain performance to the existing methods. Finally, we showcase that FG-Net is a data-efficient approach. With only 100 training samples, it can achieve decent performance.

Our major contributions are as follows. (i) We propose FG-Net, a data-efficient method for generalizable facial action unit detection.
To the best of our knowledge, we are the first to utilize StyleGAN model features for AU detection. (ii) Extensive experiments on the widely-used DISFA and BP4D datasets show that FG-Net has a strong generalization ability for heatmap-based AU detection achieving superior cross-domain performance and maintaining competitive within-domain performance compared to the state-of-the-art. (iii) FG-Net is data-efficient. The performance of FG-Net trained on 1k samples is close to the whole set ($\sim$100k).

\section{Related Work}
\noindent \textbf{Facial Action Unit Detection.}
A facial action unit is an indicator of activation of an individual or a group of muscles, \eg, cheek raiser (AU6). AUs are formalized by Paul Ekman in Facial Action Coding System (FACS) \cite{ekman1977facial}. Previous studies explore attention mechanism \cite{shao2021jaa, tang2021piap, jacob2021facial} or self-supervised learning \cite{chang2022knowledge} to get discriminative representations for AU detection.

% Zhao \etal \cite{zhao2016deep} propose Deep Region and Multi-label Learning (DRML). DRML is trained with region learning (RL) and multi-label learning (ML) and is able to identify more specific regions for different AUs than conventional patch-based methods.
Shao \etal \cite{shao2021jaa} propose J$\hat{\text{A}}$A-Net for joint AU detection and face alignment. J$\hat{\text{A}}$A-Net uses adaptive attention learning to refine the attention map for each AU.
Tang \etal \cite{tang2021piap} propose a joint strategy called PIAP-DF for AU representation learning. PIAP-DF involves pixel-level attention for each AU and individual feature elimination and utilizes the unlabeled data to mitigate the negative impacts of wrong labels.
Jacob \etal \cite{jacob2021facial} combine transformer-based architectures with region of interest (ROI) attention module, per-AU embeddings, and correlation module to capture relationships between different AUs.
Chang \etal \cite{chang2022knowledge} propose a knowledge-driven self-supervised representation learning framework. AU labeling rules are leveraged to design facial partition manners and determine correlations between facial regions.

Recent work on AU detection use graph neural networks \cite{zhang2020region, song2021hybrid, luo2022learning}. 
Zhang \etal \cite{zhang2020region} utilize a heatmap regression-based approach for AU detection. The ground-truth heatmaps are defined based on the ROI for each AU. Besides, the authors utilize graph convolution for feature refinement.
Song \etal \cite{song2021hybrid} propose a hybrid message-passing neural network with performance-driven structures (HMP-PS). A performance-driven Monte Carlo Markov Chain sampling method is proposed for generating the graph structures. Besides, hybrid message passing is proposed to combine different types of messages.
Luo \etal \cite{luo2022learning} propose an AU relationship modeling approach that learns a unique graph to explicitly describe the relationship between each pair of AUs of the target facial display.
Previous studies on AU detection achieve promising within-domain performance. However, the generalization ability, \ie, cross-domain performance, for AU detection has not been widely investigated.

Ertugrul \etal \cite{ertugrul2019cross, ertugrul2020crossing} demonstrate that the deep-learning-based AU detectors achieve poor cross-domain performance due to the variations in the cameras, environments, and subjects.
Tu \etal \cite{tu2019idennet} propose Identity-Aware Facial Action Unit Detection (IdenNet). IdenNet is jointly trained by AU detection and face clustering datasets that contain numerous subjects to improve the model's generalization ability.
Yin \etal \cite{yin2021self} propose to use domain adaptation and self-supervised patch localization to improve the cross-corpora performance for AU detection. However, this method requires data from the target domain for domain adaptation.
Hernandez \etal \cite{hernandez2021deepfn} conduct an in-depth analysis of performance differences across subjects, genders, skin types, and databases. To address this gap, they propose deep face normalization (DeepFN) that transfers the facial expressions of different people onto a common facial template.

In this paper, without using any data from the target domain, we improve the cross-corpus AU detection with the semantic-rich features from a generative model trained on a large-scale and diverse dataset.

\noindent \textbf{Face Understanding with Generative Models.}
Generative models provide an estimate of the distribution of training samples \cite{bond2021deep}. Prior work utilizing generative models for face understanding has mainly focused on semantic segmentation \cite{zhang2021datasetgan, baranchuk2021label, li2021semantic} and landmark detection \cite{zhang2021datasetgan, xu2021generative}.

Zhang \etal \cite{zhang2021datasetgan} introduce DatasetGAN, an automatic procedure to generate massive datasets of high-quality semantically segmented images requiring minimal human effort. The authors show how the GAN latent code can be decoded to produce a semantic segmentation of the image and allow the decoder to be trained with only a few labeled examples.
Baranchuk \etal \cite{baranchuk2021label} demonstrate that feature maps from diffusion models \cite{dhariwal2021diffusion} can capture the semantic information and appear to be excellent pixel-wise representations.
Li \etal \cite{li2021semantic} propose semanticGAN, a generative adversarial network that captures the joint image-label distribution. The proposed semanticGAN showcases an extreme out-of-domain generalization ability, such as transferring from real faces to paintings, sculptures, and even cartoons and animal faces.
Xu \etal \cite{xu2021generative} consider the pre-trained StyleGAN generator as a learned loss function and train a hierarchical encoder to get visual representations, namely GH-Feat, for input images. GH-Feat has strong transferability to both generative and discriminative tasks.

Previous studies show that the hidden states from the generative models are powerful representations for face understanding. However, to the best of our knowledge, no existing work adapts such architectures to AU detection.
Zhang \etal \cite{zhang2021datasetgan} and Baranchuk \etal \cite{baranchuk2021label} extract pixel-wise features and treat each pixel as a training sample, leading to extreme inefficiency due to the per-sample computational overhead. More importantly, inference with singe-pixel features lacks the inductive bias (local features), crucial to AU detection shown in the previous studies \cite{zhao2016deep, shao2021jaa}. In addition, semanticGAN \cite{li2021semantic} has to encode the input image to the latent space in an optimization-based manner for inference, which is extremely time-consuming. Thus semanticGAN can be only tested with a few samples. The limitation of this approach does not allow for training or testing with larger datasets. In this paper, we propose a Pyramid CNN Interpreter to detect the heatmaps, representing activated AUs, for AU detection, which is more efficient and can capture both local and global information.
GH-Feat \cite{xu2021generative} is the closest method to ours that extracts latent code representations from generative models. The major differences between GH-Feat and our method are: (i) GH-Feat extracts the 1D latent code features while FG-Net further decodes the latent codes to images and gets the 2D feature maps. (ii) GH-Feat is trained in a multi-stage manner while our method is end-to-end. GH-Feat utilizes the StyleGAN generator as a learned loss function and trains a hierarchical encoder and then uses this encoder to extract visual representations for downstream tasks. The whole pipeline requires more than 700 GPU hours due to the complicated training process while FG-Net only needs 10 GPU hours for training.

\begin{figure}[t]
    \centering
    \includegraphics[trim=0 350 570 0, clip, width=0.3\textwidth]{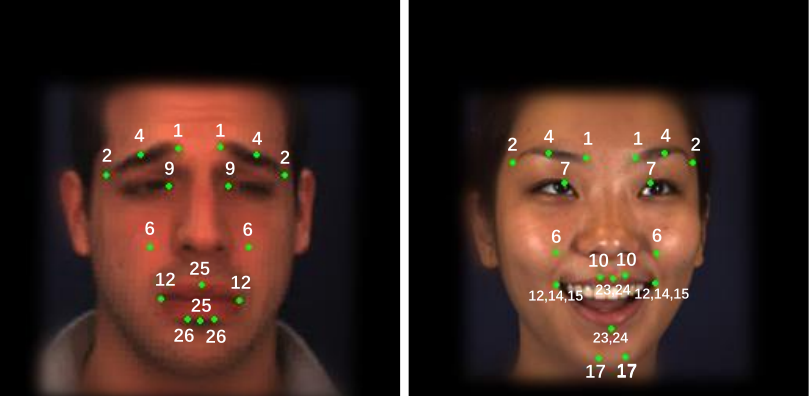}
    \caption{Visualizations of the ROI centers for DISFA (left) and BP4D (right). AU indices are labeled above or below.}
    \label{fig:roi}
\end{figure}

\begin{figure*}[t]
    \centering
    \includegraphics[trim=40 210 105 0, clip, width=0.9\textwidth]{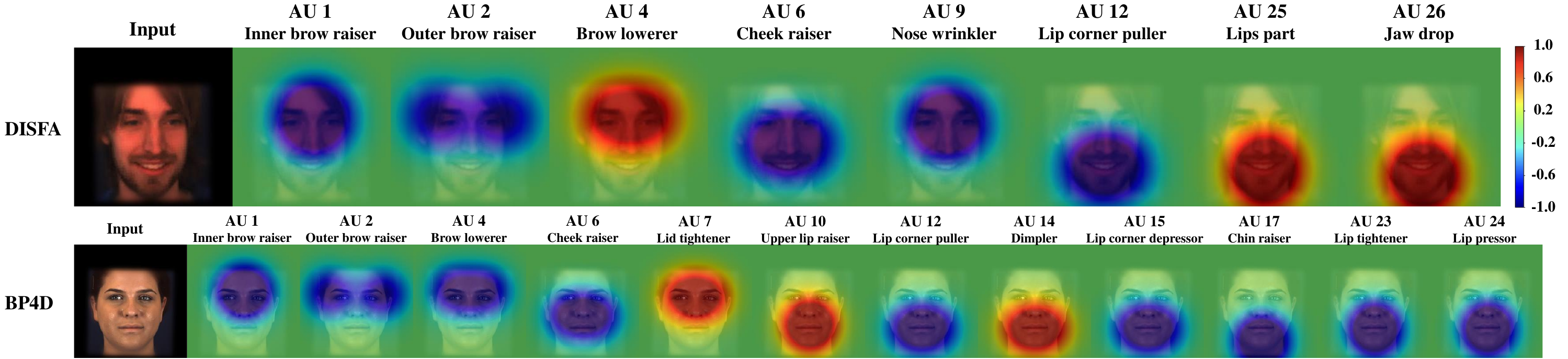}
    \caption{Visualization of the generated ground-truth heatmaps on DISFA (first row) and BP4D (second row). We generate one heatmap for every AU which has two Gaussian windows with the maximum values at the two ROI centers (see Figure \ref{fig:roi}). The peak value is either $1$ (red, AU is active) or $-1$ (blue, AU is inactive).}
    \label{fig:viz}
\end{figure*}

\section{Methods}
\subsection{Problem Formulation}
\noindent \textbf{Facial Action Unit Detection.} Given a video set $S$, for each frame $x \in S$, the goal is to detect the occurrence for each AU $a_i$ $(i=1, 2, ..., n)$ using function $\text{F}(\cdot)$.

\begin{equation}
    a_1, a_2, ..., a_n = \text{F}(x),
\end{equation}
\noindent where $n$ is the number of AUs. $a_i = 1$ if the AU is active otherwise $a_i = 0$.

\subsection{Overview}
Figure \ref{fig:model} illustrates an overview of the proposed FG-Net. FG-Net first encodes and decodes the input image with the pSp encoder \cite{richardson2021encoding} and the StyleGAN2 generator \cite{karras2020analyzing} pre-trained on the FFHQ dataset \cite{karras2019style}.

During the decoding, FG-Net extracts feature maps from the generator. Leveraging the features extracted from a generative model trained on a large-scale and diverse dataset, FG-Net offers a higher generalizability for AU detection.

To take advantage of the pixel-wise representations from the generator, FG-Net is designed to detect the AUs using a heatmap regression. To keep the training efficient and capture both local and global information, a Pyramid-CNN Interpreter is proposed to incorporate the multi-resolution feature maps in a hierarchical manner and detect the heatmaps representing facial action units.

\subsection{Model}
\noindent \textbf{Prerequisites.}
Our proposed method is built on top of the StyleGAN2 generator \cite{karras2020analyzing}. The StyleGAN2 generator decodes a latent code $z \in \mathcal{Z}$ sampled from $\mathcal{N}(0, I)$ to an image. The latent code $z$ is first mapped to a style code $w \in \mathcal{W}$ by a mapping function. Both $z$ and $w$ have $512$ dimensions. There are $k$ synthesis blocks (in practice $k=9$) and each block has two convolution layers and one upsampling layer. Each convolution layer is followed by an adaptive instance normalization (AdaIN) layer \cite{huang2017arbitrary} which is controlled by the style code $w$.
However, for image inversion which encodes the images into the latent space, $\mathcal{W}$-space has limited expressiveness and thus can not fully reconstruct the input \cite{xia2022gan}. Therefore, prior works \cite{abdal2019image2stylegan, richardson2021encoding} extend $\mathcal{W}$-space to $\mathcal{W}^+$-space where a different style code $w$ is fed to each AdaIN layer. $\mathcal{W}^+$-space alleviates the reconstruction distortion. The dimension of $w^+ \in \mathcal{W}^+$ is $18\times 512$.

\noindent \textbf{Feature Extraction from StyleGAN2.}
To extract features from the StyleGAN2 generator, we first encode the input image to the latent space and then decode the latent code.

Prior work \cite{li2021semantic} encodes the input image in an optimization-based manner. Optimization-based methods iteratively optimize a reconstruction objective which is extremely time-consuming. Instead, we utilize the pSp encoder $\text{E}$ \cite{richardson2021encoding} to encode the input image $x \in \mathcal{X}$ and get the latent code $w^+ \in \mathcal{W}^+$ via $w^+ = \text{E}(x)$.

Despite the efficient encoding of the pSp encoder, the generator features may not capture the key facial features for AU detection. To address the problem, we fine-tune the encoder and the generator during training.
Then, we decode the latent code with the StyleGAN2 generator $\text{G}$ \cite{karras2020analyzing} to obtain image $x^\prime = \text{G}(w^+)$.
During decoding, we extract the intermediate activations from the generator. To keep the training efficient, unlike the previous work \cite{zhang2021datasetgan} which extracts the outputs from all the AdaIN layers \cite{huang2017arbitrary}, we only extract the hidden states after the second AdaIN layer in each block.
We denote the feature maps we get from the $k$ blocks as $\{f_1, f_2, ..., f_{k}\} = \text{G}^\prime(w^+) = \text{G}^\prime(\text{E}(x))$.

\noindent \textbf{Heatmap Detection.}
The proposed method detects the AU occurrences in a heatmap regression-based approach. We generate the ground-truth heatmaps following the previous work \cite{zhang2020region}. We first define the Region of Interest (ROI) for each AU. We select two points on the face based on the most representative landmarks (see Figure \ref{fig:roi}, detailed positions are provided in the supplementary).

Then, we generate the ground-truth heatmaps with the definition of ROI. Figure \ref{fig:viz} gives the visualization of the ground-truth heatmaps on DISFA and BP4D. Formally, given a face image $x \in \mathbb{R}^{w \times h \times 3}$, we generate $n$ ground-truth heatmaps $m_1, m_2, ..., m_n \in \mathbb{R}^{w \times h}$ with the AU labels, where $n$ is the number of AUs. Specifically, for heatmap $m_i$, we add two Gaussian windows $g_{i}^1$ and $g_{i}^2$ with the maximum value at the two ROI centers $c_{i}^1$ and $c_{i}^2$ following \cite{zhang2020region}.
\begin{equation}
    g_i^j(p) = \lambda_i
 \exp(-\frac{||p-c_{i}^j||_2^2}{2\sigma^2}), \quad j=1,2,
 \label{eq:roi}
\end{equation}
\begin{equation}
    m_i(p) = g_i^1(p)+g_i^2(p).
\end{equation}
\noindent where $p$ is the pixel location ($p \in [1,w] \times [1,h]$). $\lambda_i$ is the indicator denoting whether the $i$-th AU is active. $\lambda_i=1$ if $a_i=1$ otherwise $\lambda_i=-1$.
$\sigma$ is the standard derivation.
We clip the heatmaps into the range of $[-1, 1]$ to make sure the peak value is either $1$ or $-1$.

After feature extraction from the generative models, prior work \cite{zhang2021datasetgan, baranchuk2021label} upsamples the features to the input resolution and concatenates them according to the channel dimension. Then, each pixel is treated as a training sample and a multi-layer perceptron (MLP) is trained to detect the semantic class. Simply upsampling and concatenating all the feature maps results in redundant and high dimensional features (in practice the number of channels is $6080$), thus leading to inefficient training and inference. More importantly, using single-pixel features for inference lacks the spatial context from nearby regions, which is crucial to AU detection, as demonstrated in the previous studies \cite{zhao2016deep, shao2021jaa}.

To address these problems, we propose a multi-scale Pyramid-CNN Interpreter $\text{H}$ for heatmap-based AU detection which incorporates the multi-resolution feature maps in a hierarchical manner (see Figure \ref{fig:model}).
Specifically, the Pyramid-CNN Interpreter $\text{H}$ contains $k$ pyramid levels, where $k$ is the number of feature maps extracted from the generator. In each pyramid level, the hidden states from the last pyramid level $c_{i-1}$ are first summed with the feature map from the generator $f_i$ and then passed through an interpreter block $\text{C}_i$. Each interpreter block consists of one Interpolate, one Convolution, one ReLU, and one BatchNorm layer. $m = c_k$ is the ultimate AU heatmap. specifically,
\begin{equation}
    c_0 = 0, c_i = \text{C}_i(c_{i-1}+f_i), i=1,2,...,k,
\end{equation}
\begin{equation}
    m = c_k = \text{H}(f_1, f_2, ..., f_k).
\end{equation}

\subsection{Training and Inference}
\noindent \textbf{Training.} The learning objective is the Mean Squared Error (MSE) loss between the ground-truth heatmap $m$ and the detected heatmap $\hat{m}$: $\mathcal{L}=||m - \hat{m}||_2^2$.

\noindent \textbf{Inference.} For each detected heatmap $\hat{m}_i$, we sum up the whole heatmap. If the sum is greater than $0$, the corresponding AU is active otherwise the AU is inactive.

\section{Experiments and Discussions}
\subsection{Datasets}
We select two publicly available datasets, \ie, DISFA \cite{mavadati2013disfa} and BP4D \cite{zhang2014bp4d}. These two datasets are widely used for AU detection and are captured from different subjects with different backgrounds and lighting conditions.

\textbf{DISFA} \cite{mavadati2013disfa} includes videos from 27 subjects, with around 130,000 frames. Each frame has labels for eight AU intensities (1, 2, 4, 6, 9, 12, 25, and 26). Following the settings of previous studies \cite{zhao2016deep, shao2021jaa, luo2022learning}, we map the AU intensity greater than 1 to the positive class.

\textbf{BP4D} \cite{zhang2014bp4d} consists of videos from 41 subjects with around 146,000 frames. Each frame has labels for 12 AU occurrences (1, 2, 4, 6, 7, 10, 12, 14, 15, 17, 23, and 24).

% \textbf{GFT} \cite{girard2017sayette} contains 96 subjects from 32 three-person groups with unscripted social interaction. Video frames are annotated for 12 AUs (1, 2, 4, 6, 7, 10, 12, 14, 15, 17, 23, and 24) which are the same as BP4D.

\begin{table}[t]
\footnotesize
\centering
    \caption{Within-domain evaluation in terms of F1 score ($\uparrow$). Except for GH-Feat and ME-GraphAU + FFHQ pre-train, all the baseline numbers are from the original papers. Our method achieves competitive performance compared to the state-of-the-art.
    }
    \begin{tabular}{l|cc}
    \toprule
    \rowcolor{Gray}
    Methods & DISFA & BP4D \\
    \midrule
    DRML \cite{zhao2016deep} & 26.7 & 48.3 \\
    IdenNet \cite{tu2019idennet} & 52.6 & 59.3 \\
    SRERL \cite{li2019semantic} & 55.9 & 62.9 \\
    UGN-B \cite{song2021uncertain} & 60.0 & 63.3 \\
    HMP-PS \cite{song2021hybrid} & 61.0 & 63.4 \\
    FAT \cite{jacob2021facial} & 61.5 & 64.2 \\
    Zhang \etal \cite{zhang2020region} & 62.0 & 63.5 \\
    J$\hat{\text{A}}$A-Net \cite{shao2021jaa} & 63.5 & 62.4 \\
    PIAP \cite{tang2021piap} & 63.8 & 64.1 \\
    Chang \etal \cite{chang2022knowledge} & 64.5 & 64.5 \\
    ME-GraphAU \cite{luo2022learning} & 63.1 & \textbf{65.5} \\
    \midrule
    ME-GraphAU + FFHQ pre-train & 59.5 & 61.1 \\
    GH-Feat \cite{xu2021generative} & 36.9 & 56.7 \\
    \midrule
    \textbf{Ours} & \textbf{65.4} & 64.3 \\
    \bottomrule
    \end{tabular}
    \label{tab:in}
\end{table}

\begin{table*}[t]
\caption{Cross-domain evaluation between DISFA and BP4D in terms of F1 scores ($\uparrow$). Our model achieves superior performance compared to the baselines. $^*$ The numbers are reported from the original paper.
}
\footnotesize
\centering
\scalebox{0.9}{\begin{tabular}{l|ccccc|a|ccccc|a}
\toprule
\rowcolor{Gray}
Direction & \multicolumn{6}{c|}{DISFA $\rightarrow$ BP4D} & \multicolumn{6}{c}{BP4D $\rightarrow$ DISFA} \\
\midrule
\rowcolor{Gray}
AU & 1 & 2 & 4 & 6 & 12 & \textbf{Avg.} & 1 & 2 & 4 & 6 & 12 & \textbf{Avg.} \\
\midrule
DRML \cite{zhao2016deep} & 19.4 & 16.9 & 22.4 & 58.0 & 64.5 & 36.3 & 10.4 & 7.0 & 16.9 & 14.4 & 22.0 & 14.1 \\
J$\hat{\text{A}}$A-Net \cite{shao2021jaa} & 10.9 & 6.7 & \textbf{42.4} & 52.9 & 68.3 & 36.2 & 12.5 & 13.2 & 27.6 & 19.2 & 46.7 & 23.8 \\
ME-GraphAU \cite{luo2022learning} & 36.5 & 30.3 & 35.8 & 48.8 & 62.2 & 42.7 & 43.3 & 22.5 & 41.7 & 23.0 & 34.9 & 33.1 \\
ME-GraphAU + FFHQ pre-train & 20.1 & 32.9 & 38.0 & \textbf{64.0} & 73.0 & 45.6 & 51.2 & 14.4 & \textbf{54.4} & 17.7 & 30.6 & 33.7 \\
GH-Feat \cite{xu2021generative} & 29.4 & 30.0 & 37.1 & \textbf{64.0} & \textbf{73.5} & 46.8 & 18.9 & 15.2 & 27.5 & \textbf{52.7} & 50.1 & 32.9 \\
\midrule
Patch-MCD$^*$ \cite{yin2021self} & - & - & - & - & - & - & 34.3 & 16.6 & 52.1 & 33.5 & 50.4 & 37.4 \\
IdenNet$^*$ \cite{tu2019idennet} & - & - & - & - & - & - & 20.1 & 25.5 & 37.3 & 49.6 & \textbf{66.1} & 39.7 \\
\midrule
\textbf{Ours} & \textbf{51.4} & \textbf{46.0} & 36.0 & 49.6 & 61.8 & \textbf{49.0} & \textbf{61.3} & \textbf{70.5} & 36.3 & 42.2 & 61.5 & \textbf{54.4} \\
\bottomrule
\end{tabular}}
\label{tab:cross}
\end{table*}

We use dlib \cite{dlib09} to detect the 68 facial landmarks for all the frames and FFHQ-alignment to align them. The detected landmarks are also used for generating the ground-truth heatmaps (see Figure \ref{fig:viz}).

\subsection{Implementation and Training Details}
All methods are implemented in PyTorch \cite{paszke17}. Code and model weights are available, for the sake of reproducibility.\footnote{https://github.com/ihp-lab/FG-Net} We use a machine with two Intel(R) Xeon(R) Gold 5218 (2.30GHz) CPUs with eight NVIDIA Quadro RTX8000 GPUs for all the experiments.
Each image is resized into $128\times128$. We train the proposed model with the AdamW optimizer \cite{loshchilov2017decoupled} for 15 epochs with a batch size of $8$ on a single GPU. The learning rate is $5e-5$. The weight decay is $5e-4$. The gradient clipping is set to $0.1$. $\sigma$ for the heatmaps (Equation \ref{eq:roi}) is $20.0$. The dropout rate is $0.1$.

\subsection{Experimental Results}
The models are evaluated for within-domain and cross-domain performance in addition to data efficiency. Cross-domain evaluation enables us to measure the generalization ability of our AU detection method. For all the experiments, F1 score ($\uparrow$) is the evaluation metric.

\noindent \textbf{Within-domain Evaluation.}
We perform within-domain evaluation on widely used DISFA and BP4D datasets. We follow the same evaluation protocols as the previous studies \cite{shao2021jaa, song2021hybrid, luo2022learning}. Both datasets are evaluated with subject-independent 3-fold cross-validation. We use two folds for training and one fold for validation and iterate three times.
We compare FG-Net to the state-of-the-art AU detection methods, including DRML \cite{zhao2016deep}, IdenNet \cite{tu2019idennet}, SRERL \cite{li2019semantic}, UGN-B \cite{song2021uncertain}, HMP-PS \cite{song2021hybrid}, FAT \cite{jacob2021facial}, Zhang \etal \cite{zhang2020region}, J$\hat{\text{A}}$A-Net \cite{shao2021jaa}, PIAP \cite{tang2021piap}, Chang \etal \cite{chang2022knowledge}, and ME-GraphAU \cite{luo2022learning}. These baseline numbers are reported from the original papers.

Previous methods do not use the FFHQ dataset for training. Thus, to make the comparison fair, two baselines are implemented and compared, \eg, ME-GraphAU + FFHQ pre-train and GH-Feat \cite{xu2021generative}. Specifically, we first pre-train the ME-GraphAU's backbones (ResNet and Swin Transformer) with the FFHQ dataset and its facial expression labels. Then we train the ME-GraphAU with the pre-trained backbones. GH-Feat extracts features from generative models and it is pre-trained on the FFHQ in a self-supervised manner. For both baselines, we implement with the officially released source codes.

Table \ref{tab:in} reports the within-domain results regarding the average performance of AUs. We provide detailed results for every individual AU in the supplementary material. On DISFA, FG-Net outperforms all the baseline methods and achieves an average F1 score of 65.4. The major improvement comes from AU1 and AU2. On BP4D, FG-Net achieves competitive performance. These results demonstrate that the pixel-wise features extracted from StyleGAN2 are beneficial for heatmap-based AU detection.

\noindent \textbf{Cross-domain Evaluation.}
We perform two directions of cross-domain evaluation, \ie, DISFA to BP4D and BP4D to DISFA. For each direction, we use two folds and one fold of data from the source domain as the training and validation set and use the target data as the testing set.
We compare the proposed method with DRML~\cite{zhao2016deep}, J$\hat{\text{A}}$A-Net~\cite{shao2021jaa}, ME-GraphAU~\cite{luo2022learning}, ME-GraphAU + FFHQ pre-train, and GH-Feat~\cite{xu2021generative} since they are open-source, and we can use the officially released source codes and model weights to conduct the experiments. In addition, we compare with Patch-MCD \cite{yin2021self} and IdenNet \cite{tu2019idennet}. The numbers are reported from the original paper. Ertugrul \etal \cite{ertugrul2019cross, ertugrul2020crossing} and Hernandez \etal \cite{hernandez2021deepfn} do not report F1 scores for the cross-domain performance of the aforementioned directions.

We report the cross-domain results in Table \ref{tab:cross}. As expected, compared to the within-domain performance, all the baseline methods suffer a considerable performance loss when evaluated across corpora. In particular, when evaluated from BP4D to DISFA, the baseline methods' performance (average F1) drops by more than 30\%, which demonstrates the challenging nature of cross-domain AU detection and the importance of developing generalizable AU detection.

Compared with DRML and J$\hat{\text{A}}$A-Net, ME-GraphAU achieves higher cross-domain performance. We suspect it is because it utilizes the pre-trained models (ResNet \cite{he2016deep} and Swin Transformer \cite{liu2021swin}) as the backbones. In addition, when we continue pre-training ME-GraphAU with the FFHQ dataset, we observe a further performance boost in both directions of cross-domain evaluations. Similarly, GH-Feat, which is trained on the FFHQ dataset, also obtains superior performance than DRML and J$\hat{\text{A}}$A-Net. The experimental results show the effectiveness of pre-training on the FFHQ dataset since it is a large and diverse facial image dataset.
Moreover, Patch-MCD utilizes unsupervised domain adaptation with unlabeled target data while IdenNet is jointly trained by AU detection and face cluster datasets (CelebA \cite{liu2015faceattributes}). Thus, with additional face data, these two methods have better cross-domain performance than the aforementioned baselines.

For both directions of cross-domain evaluation, our proposed method achieves superior performance compared to the baselines. Specifically, when evaluated from BP4D to DISFA, FG-Net can outperform the baselines by 15\% in terms of the average F1 score. The major improvement comes from AU1 and AU2, which is consistent with the findings in within-domain evaluation. Overall, the results showcase that features extracted from the StyleGAN2 generator are generalizable and thus improve the performance for cross-domain AU detection, showing its potential to solve AU detection in a real-life scenario.

\begin{figure}[t]
    \centering
    \includegraphics[trim=0 220 245 0, clip, width=0.35\textwidth]{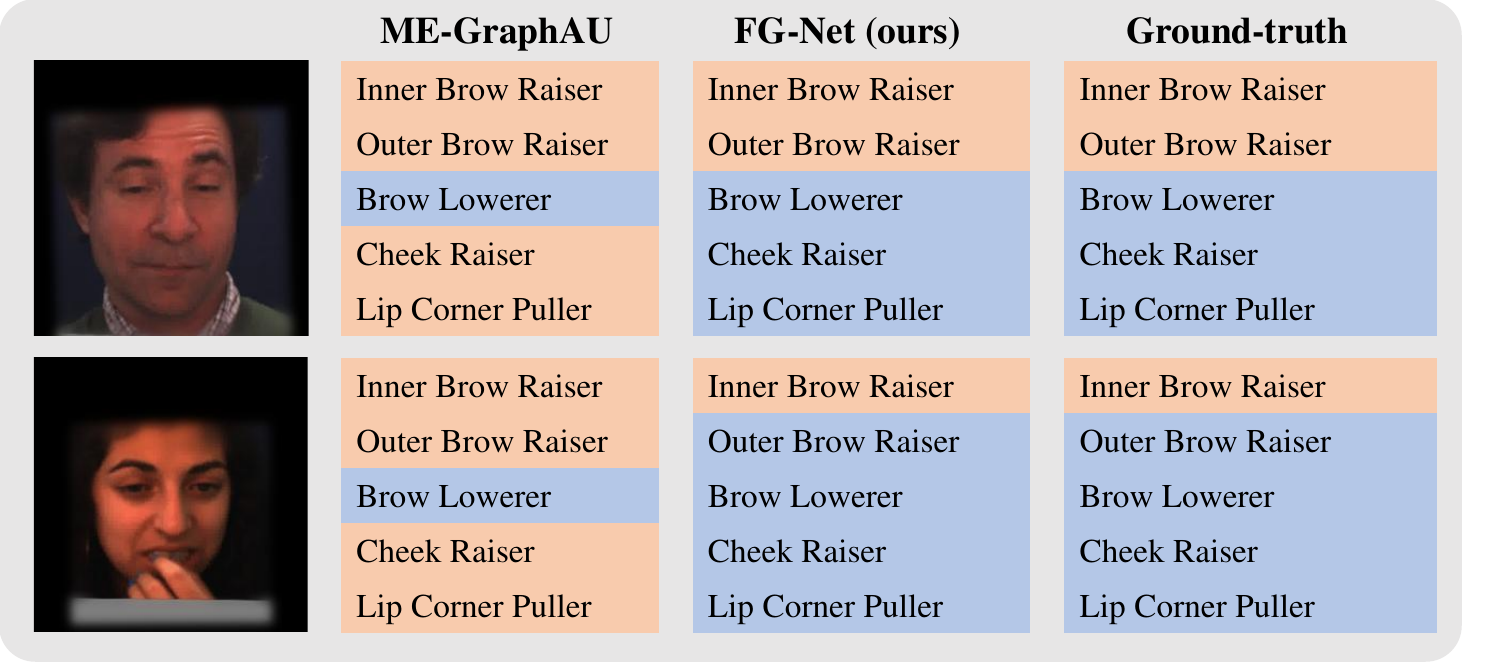}
    \caption{Case analysis on ME-GraphAU \cite{luo2022learning} and FG-Net. Models are trained on BP4D and tested on DISFA. \textcolor{orange}{Orange} means active AU while \textcolor{blue}{blue} means inactive AU. FG-Net is more accurate than ME-GraphAU.}
    \label{fig:comparison}
\end{figure}

We present two qualitative examples of cross-domain prediction in Figure~\ref{fig:comparison}. The models are trained on the BP4D dataset and tested on the DISFA dataset. ME-GraphAU fails in those two cases while the proposed FG-Net method accurately predicts the action units.

\begin{figure}[t]
\centering
    \begin{subfigure}{0.21\textwidth}
    \centering
    \includegraphics[width=\textwidth]{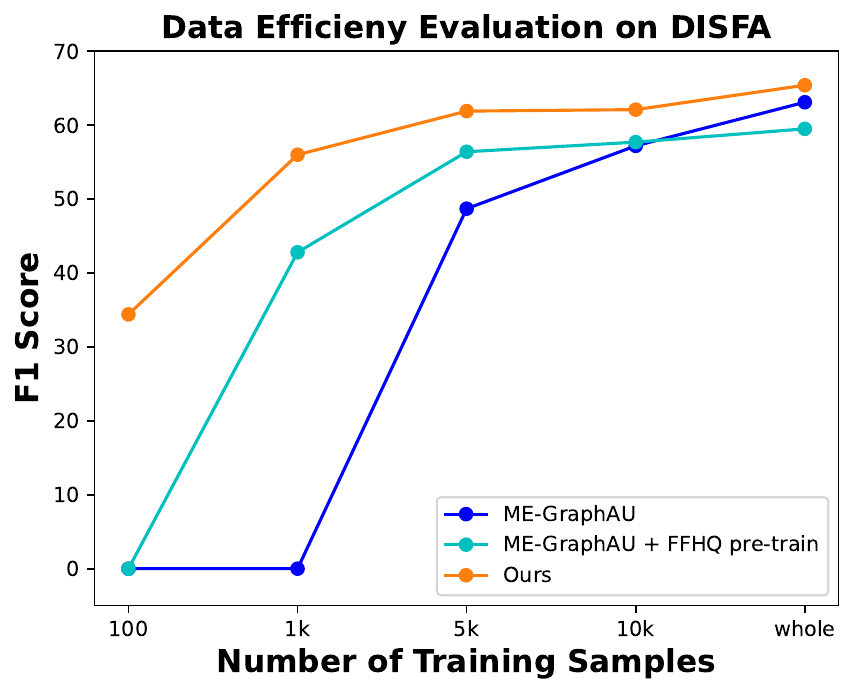}
    \caption{Evaluation on DISFA.}
    \label{fig:efficiency_disfa}
    \end{subfigure}
    \begin{subfigure}{0.21\textwidth}
    \centering
    \includegraphics[width=\textwidth]{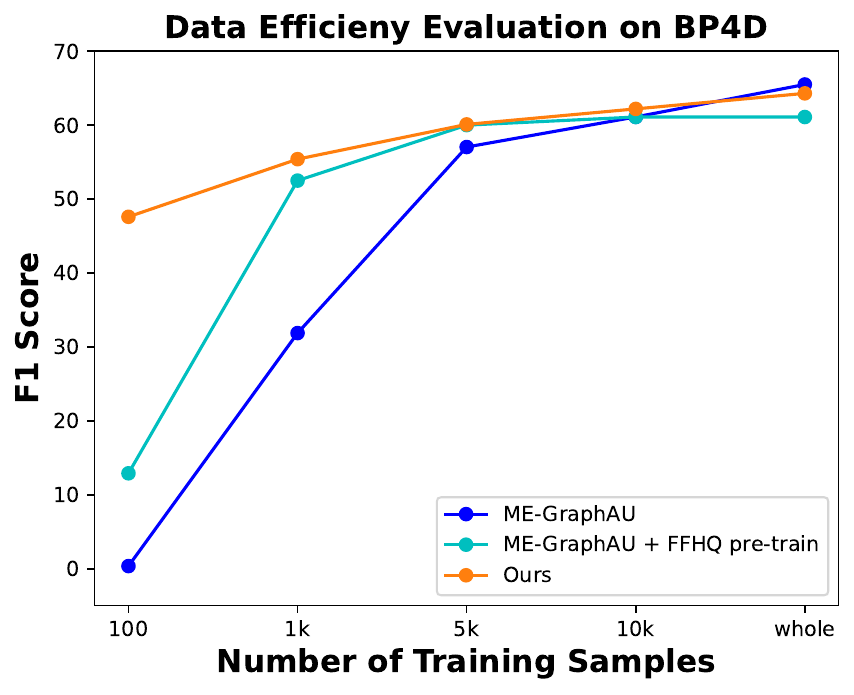}
    \caption{Evaluation on BP4D.}
    \label{fig:efficiency_bp4d}
    \end{subfigure}
    \caption{Data efficiency evaluation with different numbers of samples. Our method is data-efficient and its performance trained on 1k samples is close to the whole set.}
    \label{fig:efficiency}
\end{figure}

\noindent \textbf{Data Efficiency Evaluation.}
To further evaluate the generalization capacity of the proposed approach, an investigation of its learning capability with limited samples is conducted through within-domain evaluation. In this evaluation, a subset of the training data is randomly selected, while the testing data remains unchanged to facilitate assessment. The model is trained using four different sample sizes: 100, 1k, 5k, and 10k. A comparative analysis is performed between our method and two other approaches, namely ME-GraphAU \cite{luo2022learning} and ME-GraphAU + FFHQ pre-train. It is noteworthy that ME-GraphAU + FFHQ pre-train and our method employ the same pre-training dataset.

The efficiency evaluation results, depicted in Figure \ref{fig:efficiency}, demonstrate the impact of data scarcity on performance for both datasets. Notably, ME-GraphAU \cite{luo2022learning} exhibits remarkably low F1 scores when trained with 100 and 1k samples on the DISFA dataset, as well as with 100 samples on the BP4D dataset. This outcome can be attributed to the limited and sparse nature of the training set, causing ME-GraphAU to predict inactive AUs predominantly. By contrast, the performance of ME-GraphAU improves when pre-trained on the FFHQ dataset, underscoring the effectiveness of utilizing this extensive and diverse facial dataset for pre-training. However, even with 100 samples from the DISFA dataset, the performance of ME-GraphAU remains at 0.

In comparison, FG-Net outperforms ME-GraphAU + FFHQ pre-train when trained with partial training data for both datasets. Notably, FG-Net trained on 1k samples achieves performance levels approaching those of the full training set. Furthermore, even with a mere 100 training samples, FG-Net manages to achieve commendable performance. These results serve as evidence of the robust generalization ability exhibited by our proposed method when confronted with limited data.

\begin{table}[t]
\footnotesize
    \centering
    \caption{Ablation study for FG-Net. F1 score ($\uparrow$) is the metric. D and B stand for DISFA and BP4D. D $\rightarrow$ B means the model is trained on DISFA and tested on BP4D and similar to B $\rightarrow$ D. (i) Our method gets better performance than GH-Feat \cite{xu2021generative}. (ii) With every component, our method achieves the highest within-domain performance while removing late features gets the best cross-domain performance.}
    \scalebox{0.9}{
    \begin{tabular}{l|ccc|ccc}
    \toprule
    \rowcolor{Gray}
     & D & B & \textbf{Avg.} & D $\rightarrow$ B & B $\rightarrow$ D & \textbf{Avg.} \\
    \midrule
    % GH-Feat \cite{xu2021generative} & 36.3 & 54.0 & 45.2 & 46.8 & 32.4 & 39.6 \\
    % \midrule
    Upscale \& concat & 64.2 & 62.7 & 63.4 & 42.5 & 35.9 & 39.2 \\
    Latent code & 68.4 & 58.8 & 63.6 & 46.4 & 47.3 & 46.9 \\
    \midrule
    - Early & 68.1 & 61.7 & 64.9 & 37.9 & 47.2 & 42.6 \\
    - Middle & 67.4 & 63.1 & 65.3 & 48.5 & 38.0 & 43.3 \\
    - Late & 67.4 & 62.8 & 65.1 & \textbf{51.2} & \textbf{56.6} & \textbf{53.9} \\
    \midrule
    FG-Net & \textbf{68.9} & \textbf{63.6} & \textbf{66.3} & 49.0 & 54.4 & 51.7\\
    \bottomrule
    \end{tabular}}
    \label{tab:ablation}
\end{table}

\noindent \textbf{Ablation Study.} We conduct three ablation experiments: (i) We compare to the existing upscaling and concatenating features proposed in \cite{zhang2021datasetgan, baranchuk2021label} (upscale \& concat). (ii) We directly compare to using latent code to predict the activations of AUs (latent code). (iii) We explore the best blocks for extracting feature maps. Specifically, we divide the features extracted from the nine synthesis blocks into three groups, where each group has three feature maps, and denote them as the early, middle, and late groups. Each time, we remove one group. We perform both within- and cross-domain evaluations for the ablation study. Note that for within-domain evaluation, we use two folds for training and one fold for validation.

Table \ref{tab:ablation} shows the within- and cross-domain performance on DISFA and BP4D.
(i) We observe that FG-Net outperforms Upscale \& concat for both within- and cross-domain settings. We believe inference with singe-pixel features lacks the inductive bias, considering local features, necessary for AU detection.
(ii) FG-Net outperforms latent code for predicting AU activations for both within- and cross-domain experiments. We think using the heatmap regression allows the model to localize where the AUs occur and improves the model's capacity. In addition, compared with the 2D feature maps, the latent codes lose the semantic-rich representations.
(iii) For the contributions of different feature maps, we observe that removing any one of the feature groups lowers the within-domain performance. Surprisingly, removing late features achieves the highest cross-domain performance. We suspect it is because the late features contain more high-frequency and domain-specific information which reduces the model's generalization ability.

\begin{figure}
    \centering
    \includegraphics[trim=0 10 90 0, clip, width=0.4\textwidth]{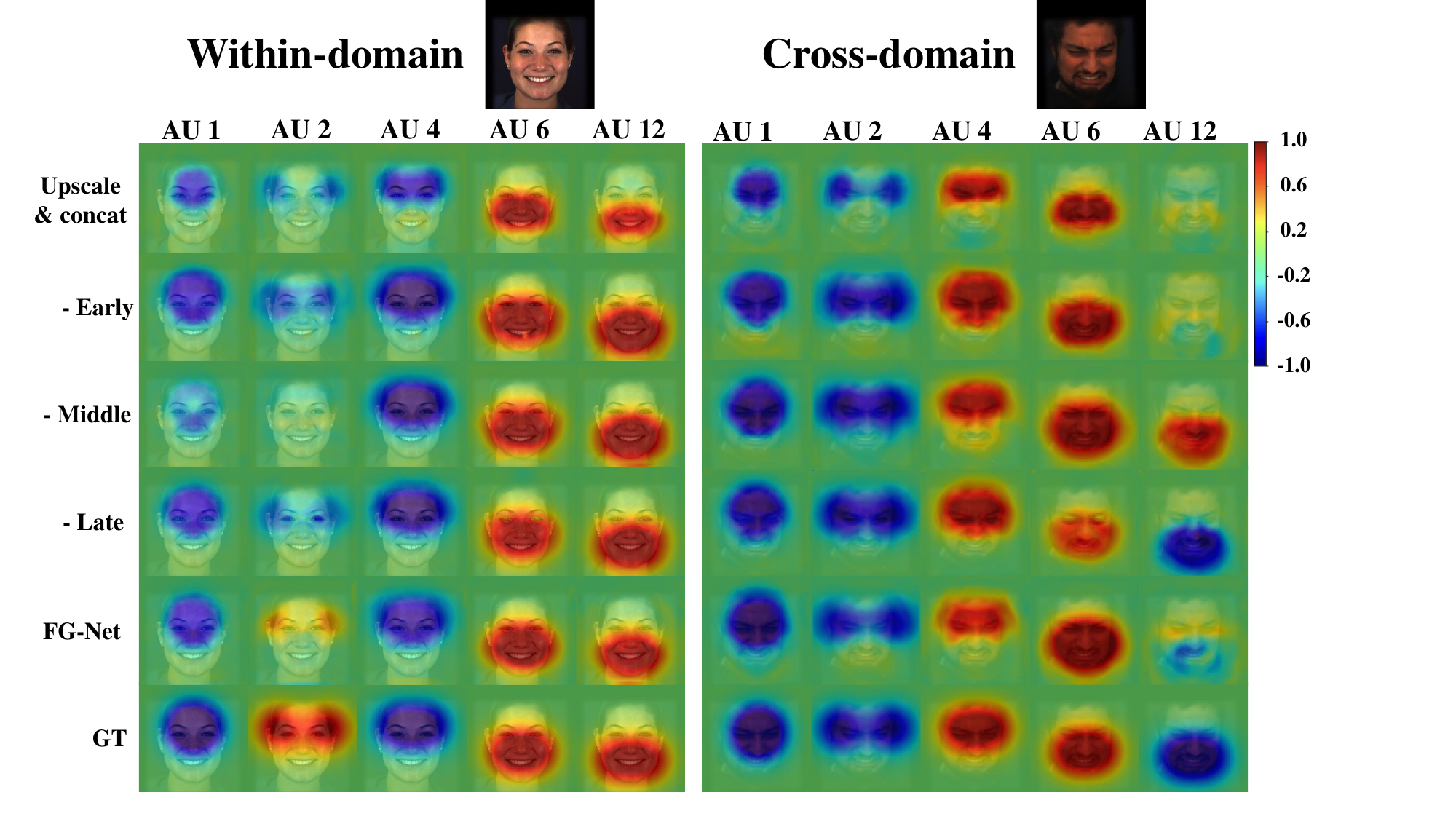}
    \caption{Visualization of the detected heatmaps for ablation study. With all the components, FG-Net detects the most similar heatmaps to the ground-truth (GT) for within-domain evaluation. Removing late features results in the best cross-domain evaluation.}
    \label{fig:ablation}
\end{figure}

We visualize the ground-truth and detected heatmaps for ablation study in Figure \ref{fig:ablation}. For the within-domain evaluation, models are trained and tested with BP4D; For the cross-domain evaluation, models are trained with BP4D and tested with DISFA. For latent code, we directly use it to predict the AU activations, thus, we do not have the detected heatmaps for latent code. For within-domain evaluation, FG-Net detects all AUs correctly, whereas the other methods output the wrong prediction for AU2 (outer brow raiser), showing that FG-Net achieves the best within-domain performance with every component. For cross-domain evaluation, both using all features and removing late features detect all AUs correctly. However, removing late features results in a more accurate heatmap for AU12 than using all features.

\subsection{Limitations}
\begin{figure}
    \centering
    \includegraphics[trim=0 100 410 0, clip, width=0.3\textwidth]{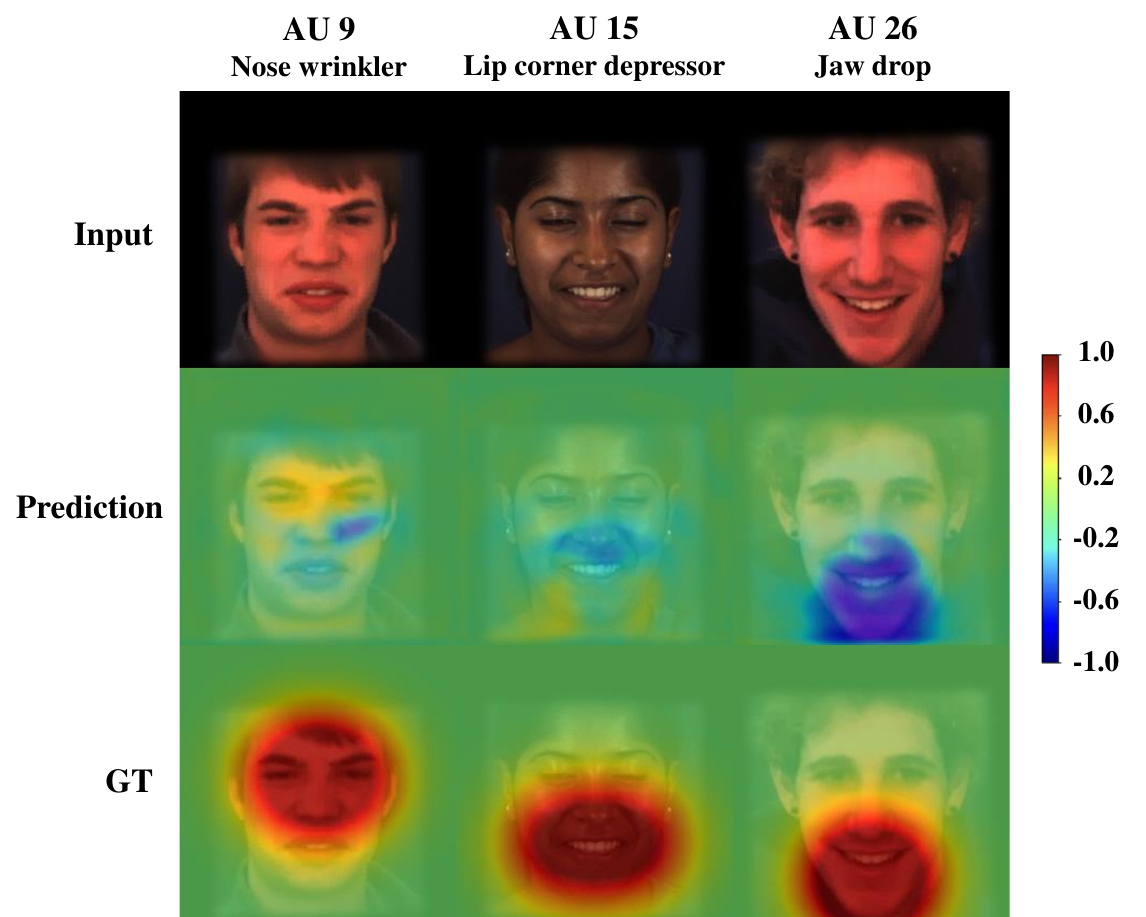}
    \caption{Visualization of the failure cases. FG-Net achieves inferior performance on AU9, AU15, and AU26.}
    \label{fig:failure}
\end{figure}

In the within-domain evaluation, FG-Net achieves inferior results on AU9 (nose wrinkler), AU15 (lip corner depressor), and AU26 (jaw drop). Failure cases are shown in Figure \ref{fig:failure}. We suspect it is because the FFHQ dataset lacks faces with such active AUs, and thus the StyleGAN2 features can not capture the corresponding information well. In addition, these failure AUs are not common in DISFA and BP4D thus they do not appear in the cross-domain evaluations and we can not evaluate the generalization for them.

FG-Net addresses the AU detection problem using a heatmap regression. Though our method can be extended to AU intensity estimation, there are only three common AUs for intensity estimation between BP4D and DISFA (6, 12, and 17) with no AU on the eyebrows. Thus, we can not properly evaluate the generalization ability of FG-Net for AU intensity estimation.

% In the ablation study, while using all feature maps achieves the best within-domain performance, its cross-domain performance is worse than using all but the late features, suggesting that using all feature maps is not optimal for generalizability. For future work, we will try automatic feature map selection for better generalization.

\section{Conclusion}
In this paper, we propose FG-Net, a data-efficient method for generalizable facial action unit detection. FG-Net extracts the generalizable and semantic-rich features from the generative model. A Pyramid CNN-Interpreter is proposed to detect AUs coded as heatmaps which makes the training efficient and captures essential information from the nearby regions. The experimental results demonstrate the challenging nature of cross-domain AU detection and the importance of developing generalizable AU detection. We show that the proposed FG-Net method has a strong generalization ability when evaluated across corpora or trained with limited data, demonstrating its strong potential to solve action unit detection in a real-life scenario.

\textbf{Social Implications.} Our work falls within the broad domain of facial expression analysis. Despite potential benefits, any surveillance technology can be misused, and sensitive private information may be revealed by malicious actors. Mitigation strategies for such misuses include restrictive licensing and government regulations.

\section{Acknowledgement}
The work of Soleymani, Yin and Chang was sponsored by the Army Research Office and was accomplished under Cooperative Agreement Number W911NF-20-2-0053. The views and conclusions contained in this document are those of the authors and should not be interpreted as representing the official policies, either expressed or implied, of the Army Research Office or the U.S. Government. The U.S. Government is authorized to reproduce and distribute reprints for Government purposes notwithstanding any copyright notation herein.

%%%%%%%%% REFERENCES
{\small
\bibliographystyle{ieee_fullname}
\bibliography{main}
}

\end{document}